%% file: example_paper.tex
\documentclass{article}
\usepackage{microtype}
\usepackage{graphicx}
\usepackage{subcaption}
\usepackage{booktabs}
\usepackage{hyperref}

\usepackage[preprint]{icml2026}
\usepackage{amsmath}
\usepackage{amssymb}
\usepackage{mathtools}
\usepackage{amsthm}
\usepackage[capitalize,noabbrev]{cleveref}
\theoremstyle{plain}

\theoremstyle{definition}

\theoremstyle{remark}

\usepackage[textsize=tiny]{todonotes}
\icmltitlerunning{Mitigating Overthinking in Large Reasoning Language Models
via Reasoning Path Deviation Monitoring}
\usepackage{enumitem}
\usepackage[table]{xcolor}
\usepackage{multirow}
\usepackage{booktabs}
\usepackage{graphicx}
\usepackage{pifont}
\usepackage{xspace}
\usepackage{algorithm}
\usepackage{algorithmic}
\usepackage{xcolor}
\usepackage{amsmath, amssymb}

\newcommand{\cmark}{\textcolor{green!60!black}{\ding{51}}}
\newcommand{\xmark}{\textcolor{red!70!black}{\ding{55}}}
\newcommand{\colorgreen}{\color{green!60!black}}

\newcommand{\OurMethod}{\textsc{RPDI-EE}\xspace}
\newcommand{\mathEasy}{\textsc{Math Easy}\xspace}
\newcommand{\mathHard}{\textsc{Math Hard}\xspace}
\newcommand{\scientificBC}{\textsc{Scientific}\xspace}
\newcommand{\userdefinedfooter}[1]{%
  \begingroup
    \renewcommand{\thefootnote}{}
    \footnotetext{\noindent\footnotesize #1}
  \endgroup
}

\begin{document}

\twocolumn[
  \icmltitle{Mitigating Overthinking in Large Reasoning Language Models\\ via Reasoning Path Deviation Monitoring}
  \icmlsetsymbol{corresponding}{\dag}
  \icmlsetsymbol{projectl}{*}
  \begin{icmlauthorlist}
    \icmlauthor{Weixin Guan}{iie,ucas}
    \icmlauthor{Liang Li}{iie,corresponding}
    \icmlauthor{Jiapeng Liu}{iie,ucas}
    \icmlauthor{Bing Li}{iie}
    \icmlauthor{Peng Fu}{iie}
    \icmlauthor{Chengyang Fang}{jxcj}
    \icmlauthor{Xiaoshuai Hao}{xiaomi,projectl}
    \icmlauthor{Can Ma}{iie}
    \icmlauthor{Weiping Wang}{iie}
  \end{icmlauthorlist}
  \icmlaffiliation{iie}{Institute of Information Engineering, Chinese Academy of Sciences, Beijing, China}
  \icmlaffiliation{ucas}{School of Cyberspace Security, University of Chinese Academy of Sciences, Beijing, China}
  \icmlaffiliation{jxcj}{School of Computer and Artificial Intelligence, Jiangxi University of Finance and Economics, Jiangxi, China}
  \icmlaffiliation{xiaomi}{Xiaomi EV}
  \icmlcorrespondingauthor{Liang Li}{liliang@iie.ac.cn}
  \icmlkeywords{Overthinking, Early Exit, Large Reasoning Language Models}
  
  \vskip 0.3in
]
\userdefinedfooter{*Project leader.}
\printAffiliationsAndNotice{}
\begin{abstract}
Large Reasoning Language Models (LRLMs) demonstrate impressive capabilities on complex tasks by utilizing long Chain-of-Thought reasoning.
However, they are prone to overthinking, which generates redundant reasoning steps that degrade both performance and efficiency.
Recently, early-exit strategies are proposed to mitigate overthinking by dynamically and adaptively terminating redundant reasoning.
However, current early-exit methods either introduce extra training overhead by relying on proxy models or limit inference throughput due to the frequent content switching between reasoning and generating probing answers.
Moreover, most early-exit methods harm LRLMs performance due to over-truncation.
Our insight stems from an observation: overthinking often causes LRLMs to deviate from the correct reasoning path, which is frequently accompanied by high-entropy transition tokens.
Given this, we propose an early-exit method deeply coupled with the native reasoning process, which leverages the path deviation index as a dedicated monitoring metric for the frequent occurrence of high-entropy transition tokens to dynamically detect and terminate overthinking trajectories. 
We conduct experiments across multiple benchmarks using LRLMs of different types and scales, and the results indicate that our method delivers the largest performance improvement over vanilla CoT compared to existing early-exit methods.
\end{abstract}
\section{Introduction}
Large Reasoning Language Models (LRLMs) \citep{guo2025deepseek, team2025qwq, yang2025qwen3} demonstrate impressive capabilities on complex tasks by leveraging long Chain-of-Thought (CoT) reasoning.
However, extended chains introduce a critical weakness: LRLMs often engage in \emph{overthinking} \citep{chen2024not, su2025between, marjanovic2025deepseek}, a phenomenon in which the model generates redundant reasoning steps that fail to meaningfully contribute to the final answer.
More seriously, overthinking degrades reasoning performance and increases reasoning latency.
Specifically, unnecessary reasoning steps increase computational costs and lead to unnecessary latency.
Meanwhile, excessive reasoning steps can distract the model during final-answer generation and lead to deviations from the correct reasoning path due to error accumulation.

\input{sources/tex/compare}

To mitigate overthinking, some researchers introduce the early-exit strategy, which truncates the reasoning process and switches directly to generating the final answer once it detects that the model has produced sufficient intermediate reasoning steps.
Early explorations \citep{muennighoff2025s1, ma2025reasoning, li2025thinkless} propose setting a fixed maximum length for reasoning to prevent the model from overthinking. 
However, this mode lacks the necessary flexibility to adapt to problems with varying complexity.
Recent studies \citep{xu2025chain, fang2025thinkless, ding2025dynamic} address this weakness by dynamically identifying early-exit points for adaptive processing of problems with varying difficulty.
Specifically, these methods divide the reasoning process into a series of reasoning segments using heuristics and evaluate whether to exit early at the boundary of each segment.

To determine early-exit points, some approaches \citep{yang2025specexit, jiang2025flashthink, zhang2025reasoning, akgul2025lynx} introduce additional proxy models as a detector.
Nevertheless, proxy models introduce extra training overhead and require specialized training to adapt them to different models and tasks.
Alternatively, other methods \citep{yang2025dynamic, yong2025think, fu2025reasoning, akgul2025lynx} identify early-exit points by probing the answer.
They generate tentative answers at reasoning segment boundaries and terminate the reasoning process once the confidence or consistency of these answers exceeds a threshold.
These approaches are more cost-effective and widely applicable, offering immediate benefits for existing large-scale models, e.g., DeepSeek-R1-671B \citep{guo2025deepseek} and Qwen3-235B \citep{yang2025qwen3}.
However, these methods often cause the generation to frequently switch between reasoning steps and tentative answers, thereby limiting the response speed.
Furthermore, we observe that some of these methods \citep{fu2025reasoning, li2025thinkless, yang2025specexit, jiang2025flashthink, zhang2025reasoning, ma2025reasoning, muennighoff2025s1} exhibit performance degradation compared to vanilla CoT in some scenarios.
This performance bottleneck might be attributed to over-truncation: LRLMs manifest spurious confidence levels, leading to suboptimal early-exit decisions and effectively 'silencing' the model before it can achieve self-rectification. 
Such observations reveal a decoupling between the consistency of tentative answers and the intrinsic quality of the model's reasoning trajectory.
Consequently, it may be necessary to \textbf{look inward} at the reasoning trajectory itself rather than solely \textbf{looking forward} at tentative answers.

Building on this `inward' perspective, this paper explores the use of latent trajectory signals to mitigate overthinking in LRLMs while safeguarding against over-truncation.
Theoretically, information entropy serves as a dynamic indicator of the model's internal uncertainty during generation. 
Moreover, existing studies suggest that high-entropy tokens within a reasoning trajectory often manifest as transition terms, such as `wait', `alternatively', or `but'. 
Intuitively, a high frequency of these transition tokens indicates that the model is producing fragmented reasoning chains, struggling to deepen a single reasoning path.
Figure~\ref{fig:intro_case} illustrates this behavior in an LRLM solving a geometry problem. 
Although the model initially finds the correct answer, it makes a small calculation error during verification. 
This error triggers a logical contradiction with the problem's conditions, leaving the model ``stuck,'' i.e., overthinking. 
This state is clearly marked by the repetitive use of transition tokens like ``Wait'' showing that the model cannot deepen the reasoning path. 
Based on this, we hypothesize that an anomalous frequency of these high-entropy tokens can serve as an internal signal of the model entering a state of overthinking.

While directly counting high-entropy tokens is intuitive, such a ``hard count'' approach relies on a specific entropy thresholds that fail to generalize across different models and task difficulties.
Through an extensive visualization of reasoning trajectories, we observe that token entropy follows a long-tail distribution: the vast majority of tokens possess negligible entropy, while a small number of transition tokens contribute most of the average entropy. 
Please refer to Section \ref{preliminary_experiment} for more details.
Inspired by these insights, we propose \OurMethod, an early-exit method based on the Reasoning Path Deviation Index (RPDI), which uses average entropy as a ``soft count'' of high-entropy tokens to avoid the instability of hard thresholds.
Specifically, \OurMethod calculates RPDI as the ratio of Local Transition Frequency (LTF) to Global Transition Frequency (GTF). 
Among them, LTF is the average entropy of the most recently generated reasoning content, reflecting the frequency of local transition tokens.
In contrast, GTF calculates the average entropy of the entire reasoning trajectory produced thus far, serving as a global baseline.
By calculating this relative frequency change, the RPDI quantifies anomalous spikes in the frequency of transition tokens relative to the overall reasoning process, allowing the system to distinguish unproductive wandering from normal thinking transitions. 
When the RPDI exceeds a predefined threshold $\lambda$, signaling that the reasoning path is deviated into overthinking, \OurMethod triggers an early exit—effectively suppressing redundant inference before it impairs performance.

Our contributions are summarized as follows:
\begin{itemize}
    \setlength{\itemsep}{0pt}
    \setlength{\parsep}{0pt}
    \setlength{\parskip}{0pt}
\item We provide a new perspective on model overthinking, identifying that it manifests internally as a surge in high-entropy transition tokens (e.g., ``Wait'', ``But'').
\item We propose \OurMethod, a novel, training-free early-exit method based on the Reasoning Path Deviation Index. 
It does not require introducing external proxy models or probing for answers, thus avoiding extra computational costs.
\item We conduct extensive experiments across multiple benchmarks using LRLMs of various types and scales.
The results demonstrate that our approach achieves the largest performance improvements over vanilla CoT and effectively mitigates the over-truncation issues prevalent in existing early-exit strategies.
\end{itemize}

\section{Related Works}

The emergence of LRLMs has demonstrated that long CoT reasoning can significantly enhance performance on complex tasks.
However, this capability also introduces \emph{overthinking}, where models produce excessively long reasoning chains that increase computational costs and may degrade performance through error accumulation and reasoning path deviation \citep{chen2024not, su2025between, gan2025rethinking}.
Existing mitigation strategies can be broadly categorized into training-time and inference-time optimizations.

\textbf{Training-time optimization} aims to encode efficient reasoning patterns directly into model parameters.
A common strategy fine-tunes models on datasets where verbose reasoning is distillation into concise chains \citep{fatemi2025concise, shen2025dast}.
Another line of work employs reinforcement learning with efficiency-driven rewards such as length penalties to encourage shorter reasoning \citep{qiao2025concise, kang2025c3ot, liu2024can}.
More radical approaches compress reasoning into latent representations \citep{yeo2025demystifying, lou2025adacot}, improving token efficiency but often at the expense of the verifiability provided by explicit reasoning.
While these methods can produce efficient reasoners, the efficiency bias becomes permanently encoded in model parameters, limiting adaptivity to dynamic computational budgets without costly retraining.

\textbf{Inference-time optimization} aims to adjust the reasoning procedure without modifying model parameters.
Multi-path reasoning generates several chains to mitigate individual errors \citep{ding2025dynamic, wang2025sampling, sun2024fast}, but is computationally expensive and subject to diminishing accuracy gains from additional reasoning paths due to high correlations across chains.
Difficulty-based approaches trigger long CoT only when problems appear challenging \citep{fang2025thinkless, jiang2025think, chuang2024learning}, but once activated, they provide little control and may still lead to overthinking.
Prompt-based techniques dynamically modulate reasoning length \citep{xu2025chain, han2024token, chen2024unlocking, renze2024benefits, lee2025well}, though they remain sensitive to prompt design and rely on the model's intrinsic ability.
Additionally, compressing generated reasoning paths can reduce context size, but does not reduce the cost of generating redundant tokens, and compressing generated reasoning paths may lead to discarding essential information \citep{yan2025inftythink, zhang2025lightthinker}.

To more precisely address the overthinking phenomenon, \textbf{early-exit strategies} have emerged as a specialized class of inference-time adaptations. 
The simplest approaches employ a Token-budget design, imposing fixed length constraints to truncate reasoning \citep{muennighoff2025s1, li2025thinkless, ma2025reasoning}.
While easy to implement, these methods lack task-level adaptivity and often lead to over-truncation, sacrificing performance for efficiency.
To improve adaptivity, some strategies utilize proxy models to monitor reasoning progress \citep{jiang2025flashthink, yang2025specexit, zhang2025reasoning, akgul2025lynx}.
Alternatively, other methods use the model's own generation as a stopping signal by interrupting the reasoning stream to probe tentative answers and determine termination points \citep{yang2025dynamic, yong2025think, fu2025reasoning, akgul2025lynx}.
Although these methods avoid expensive training, they either introduce training overhead for proxy models or suffer from substantial context-switching overhead during probing.
We compare the characteristics of various early-exit methods across multiple dimensions in a tabular format; please refer to Appendix ~\ref{compare} for details.

\input{sources/tex/high_entropy_word_weight}
\section{Preliminary Experiment}
\label{preliminary_experiment}
Before establishing our methodology, we conduct a preliminary analysis to characterize token entropy across various reasoning trajectories.
The goal is to investigate how the token entropy fluctuates as the model navigates complex logical steps and to identify statistical patterns that might govern these transitions.
Following previous work \citep{wang2025beyond}, we utilize the vanilla CoT to conduct statistical analysis on the 45.9 million tokens generated by the DeepSeek-R1-Distill-Qwen series (7B, 14B, and 32B) models across seven datasets. 
To ensure consistency, all experimental configurations are aligned to the vanilla CoT settings used in our main experiments.

As illustrated in Figure~\ref{fig:preliminary1}, we observe that the distribution of entropy contributions exhibits a striking long tail: the vast majority of tokens possess negligible entropy, while a few transition tokens contribute most to the average entropy.
Specifically, the first $60\%$ of tokens (sorted by entropy) contribute almost zero to the total entropy sum.
This implies that during the reasoning process, the model maintains extremely low entropy for most tokens, while a few high-entropy tokens predominantly drive the overall average entropy.
Furthermore, we visualize the $20\%$ of tokens that contributed most to the average entropy in Figure~\ref{fig:preliminary1}, and the results are displayed in Figure~\ref{fig:preliminary2}. 
It can be seen that the high-entropy tokens are mainly transition tokens.

In summary, we can draw an empirical insight: since the average entropy is mainly determined by these few high-entropy tokens, it can serve as a soft counting proxy for the density of transition tokens in a sequence.

\input{sources/tex/high_entropy_word}

\section{Methodology}

\input{sources/tex/method}

This paper proposes a dynamic early-exit method based on ``looking inward'' at the reasoning trajectory itself, aiming to achieve precise early-exit decisions by analyzing the internal states of the reasoning process in real time to mitigate the ``overthinking'' issue in LRLMs.
As illustrated in Figure 1, the framework consists of three components: Real-time Trajectory Entropy Tracking, Path Deviation Index Construction, and Dynamic Early-Exit.
The algorithm is detailed in Appendix A.

\subsection{Real-time Trajectory Entropy Tracking}

To provide a robust data foundation for subsequent path deviation detection, this component extracts and quantifies uncertainty signals from the model’s continuous generation process in real-time.
Based on the ``looking inward'' concept, it directly leverages the internal probability distribution during reasoning. Through lightweight computation, the component enables efficient monitoring of entropy dynamics without interfering with the normal reasoning process.

\paragraph{Token-level Entropy Extraction}
To quantify the model's uncertainty at each token generation step, we extract the complete probability distribution from the output layer and compute its Shannon entropy.
Specifically, at the $i$-th generation step, the model produces a probability distribution $p_i$ for the next token $t_i$, conditioned on the initial prompt $P$ and the previously generated reasoning content $R$.
The entropy of this distribution is defined as:
\begin{equation}
H(t_i) = -\sum_{v \in \mathcal{V}} p_i(v) \log p_i(v),
\end{equation}
where $\mathcal{V}$ denotes the model's vocabulary.
The resulting entropy values $H(t_i)$ are recorded sequentially to form an entropy sequence $\mathcal{H}$.
This process maps the model's implicit uncertainty onto a continuously measurable scalar sequence, providing a rigorous quantitative framework for identifying anomalous patterns of high-entropy tokens.

\paragraph{Incremental Entropy Accumulation}
To efficiently compute both local and global entropy sums without redundant calculations, we design an incremental entropy accumulation mechanism.
This mechanism maintains running sums of entropy values, eliminating the need to re-sum the entire history each time local or global statistics are required.
Specifically, we maintain two accumulator variables: $\mathcal{S}^i_{\text{global}}$ for the cumulative sum of all token entropies from the beginning of reasoning, and $\mathcal{S}^i_{\text{local}}$ for the sum of entropies within the most recent sliding window of size $W$.
Upon generating a new token $t_i$ with entropy $H(t_i)$, both accumulators are updated simultaneously:
\begin{equation}
\mathcal{S}^i_{\text{global}} = \mathcal{S}^{i-1}_{\text{global}} + H(t_i),
\end{equation}
\begin{equation}
\mathcal{S}^i_{\text{local}} = \mathcal{S}^{i-1}_{\text{local}} + H(t_i).
\end{equation}
To maintain the sliding window property, when the number of generated tokens $i$ exceeds the window size $W$, we subtract the entropy value that exits the window from $S_{\text{local}}$:
\begin{equation}
\mathcal{S}^i_{\text{local}} = \mathcal{S}^{i-1}_{\text{local}} - \mathcal{H}[i-W],\quad\text{if } i > W.
\end{equation}
This $O(1)$ update strategy ensures that both local and global entropy sums can be computed efficiently, avoiding the $O(n)$ complexity of repeatedly summing over the entire history or window. The stored entropy sequence $\mathcal{H}$ is only used for this subtraction operation, not for recomputing sums from scratch.

\subsection{Reasoning Path Deviation Index}
We introduce the Reasoning Path Deviation Index (RPDI) to transform the entropy sequence into a stable, interpretable indicator signal, enabling early-exit decisions that adapt flexibly to varying model capabilities and task difficulties.
RPDI is computed as the ratio of the Local Transition Frequency (LTF) to the Global Transition Frequency (GTF).

\paragraph{Local Transition Frequency}
The Local Transition Frequency (LTF) is measured as the average token entropy within the sliding window, characterizing the frequency of transition tokens in the model's recently generated content:
\begin{equation}
LTF_i = \frac{\mathcal{S}^i_{\text{local}}}{W}.
\end{equation}
A significant increase in LTF typically indicates that the model has generated a large number of transition tokens within a short period. 
This frequent switching of reasoning paths is closely associated with overthinking.

\paragraph{Global Transition Frequency}
However, a fixed LTF threshold lacks generalizability across different scenarios because the baseline frequency of transition tokens depends on a model's entropy distribution and the task's complexity.
To solve this problem, we further introduce the Global Transition Frequency (GTF) to provide an adaptive reference baseline.
GTF is defined as the average entropy of all generated tokens from the initiation of reasoning to the current point.
It quantifies the overall frequency of transition token occurrences, providing an adaptive reference baseline for detecting relative abnormal increases in this frequency:
\begin{equation}
GTF_i = \frac{\mathcal{S}^i_{\text{global}}}{i}.
\end{equation}

In the $i$-th reasoning step, the path deviation index standardizes the measurement of the abnormal increase in local uncertainty by calculating the ratio of $LTF_i$ to $GTF_i$:
\begin{equation}
RPDI_i = \frac{LTF_i}{GTF_i}.
\end{equation}
When the model advances steadily along the correct reasoning path, $LTF_i$ and $GTF_i$ are roughly comparable, and $RPDI_i$ fluctuates around $1$.
However, when the model falls into overthinking and frequently produces high-entropy content locally, $LTF_i$ becomes significantly higher than $GTF_i$, causing $RPDI_i$ to increase sharply.
Therefore, $RPDI_i$, as a dimensionless metric, can reliably indicate whether the reasoning path has substantially deviated.

\subsection{Dynamic Early-Exit}

Theoretically, during the LRLM reasoning process, when RPDI-EE detects that the RPDI score of the current tokens exceeds the threshold, it will make an early stopping decision, causing the LLM to switch from the thinking mode to the answering mode.
To achieve this within a prescribed reasoning token budget $L_{\text{max}}$, \OurMethod inserts an explicit mode termination marker $T$ (e.g., \texttt{</think>}) into the current trajectory $R$ to prompt the LRLM to conclude its reasoning and generate the final answer $A$.
Subsequently, the LRLM generates the final answer $A$ based on $P \oplus R$ within the remaining token budget ($L_{\text{max}} - i$).
This process can be formalized as:
\begin{equation}
A = \mathrm{LLM}([P \oplus R \oplus T]), \quad \text{for} \ RPDI_i > \lambda,
\end{equation}
where $\oplus$ denotes the concatenation operation, $\lambda$ is the early exit threshold, which is a pre-defined hyperparameter.

In practice, the \OurMethod is hindered by a primary challenge: computational redundancy arising from per-token RPDI evaluation. 
To address this issue, we incorporate a \textbf{Boundary-Triggered} RPDI Evaluation mechanism to maintain semantic integrity while reducing overhead.
Specifically, the \OurMethod computes the RPDI and determines whether it exceeds the threshold $\lambda$ if and only if the current token $t_i$ belongs to a predefined set of boundary symbols $\mathcal{B}$.
This design is followed by previous work~\citet{yang2025dynamic}, and we employ a boundary set $\mathcal{B}$ identical to theirs.
These symbolic boundaries typically signify the completion of a functional semantic unit.

\section{Experimental Setup}

\paragraph{Model Selection and Architectures}
To evaluate the efficacy and generalizability of \OurMethod across varying model scales and training paradigms, we conduct experiments on a diverse suite of eight open-source large reasoning language models.
Our selection encompasses the DeepSeek-R1-Distill-Qwen series of distillation models, ranging from the lightweight 1.5B and 7B variants to the larger 14B and 32B models \citep{guo2025deepseek}, as well as the DeepSeek-R1-Distill-Llama series featuring the 8B and 70B models.
To further test our method on state-of-the-art long reasoning architectures, we incorporate the Qwen3 Thinking series, specifically the Qwen3-30B-A3B-Thinking-2507 and the massive Qwen3-235B-A22B-Thinking-2507 \citep{yang2025qwen3}.
This broad spectrum of models ensures a comprehensive evaluation of \OurMethod's performance across both distillation LRLMs and those RL-based ones.

\paragraph{Benchmarks and Evaluation}
Our evaluation benchmarks cover a wide range of mathematical and scientific reasoning tasks, categorized by their complexity and domain.
For foundational and competitive mathematics we utilize GSM8K~\citep{cobbe2021training}, AMC23~\citep{amc23}, and the MATH500~\citep{hendrycks2021measuring} datasets, denoted as \textbf{\mathEasy}.
To evaluate the models on elite-level problem-solving, we include the AIME2024, AIME2025~\citep{aime}, and OlympiadBench~\citep{he2024olympiadbench}, denoted as \textbf{\mathHard}.
Finally, we assess high-level scientific reasoning using GPQA-Diamond~\citep{rein2024gpqa} (marked as \textbf{\scientificBC}), a benchmark comprising graduate-level questions verified by subject-matter experts in physics, biology, and chemistry.
This multi-level benchmark suite provides a rigorous testing ground for \OurMethod's capabilities.
For evaluation, we align with DEER~\citep{yang2025dynamic} and employ accuracy (Acc.) and generated token length (Len.) as metrics.

\paragraph{Baselines}
To establish a clear performance gain, we compare \OurMethod against five representative baseline strategies.
The first is Vanilla CoT, which represents standard autoregressive generation.
To evaluate early-exit performance across different paradigms, we include NoThinking~\citep{ma2025reasoning} and ThinkLess~\citep{li2025thinkless} as representatives of fixed token-budget methods.
For comparison with dynamic early-exit frameworks, we chose DEER~\citep{yang2025dynamic} and Dynasor-CoT~\citep{fu2025reasoning}, both of which use answer probing and context switching to determine termination.
Limited by computing resources and costs, we evaluate all baselines only on DeepSeek-R1-Distill-Qwen-7B/14B/32B and Qwen3-30B-A3B-Thinking-2507.
For the other four models, we compare only with the basic Vanilla-CoT and the most competitive DEER method.

Detailed implementation specifics, including hyperparameters, computational environment, and further configuration details, are provided in Appendix \ref{app:experimental-details}.

\input{sources/tex/main_result}
\section{Experimental Results and Analysis}
\label{sec:results_and_analysis}

\subsection{Main Results}

We conduct extensive experiments across multiple benchmarks, including \mathEasy, \mathHard, and \scientificBC, using eight models of various types and scales, including those trained with distillation and reinforcement learning.
As shown in table~\ref {tab:comprehensive_comparison_final}, experimental results indicate that \OurMethod delivers the most significant performance improvement over vanilla CoT on all tested models, with an average accuracy gain of $3.9\%$.
This demonstrates the effectiveness and adaptability of our approach across different architectures and problem types.
Furthermore, we observe that \OurMethod achieves larger gains on distillation models, with an average accuracy improvement of $5.1\%$.
We consider this is because distillation models are more prone to overthinking, as they often capture only the surface patterns of long CoT reasoning rather than deeply understanding the underlying logic \citep{dai2025capture, wang2025wait}.
Notably, while \OurMethod achieves the most significant performance gains, it yields a more modest reduction in token consumption than aggressive early-exit methods.
We consider this is because \OurMethod only terminates reasoning when the LRLM is truly "stuck," rather than prematurely stopping when it is still steadily progressing along the correct reasoning path and exhibiting spurious confidence. 
This also implies that, compared to other early-exit baselines, \OurMethod preserves the model's potential to self-correct its inference trajectory.

Overview, the above results validate that \OurMethod effectively mitigates overthinking by monitoring internal reasoning trajectory signals to detect reasoning path deviations.

\subsection{Analysis of Early-Exit Triggering Impact}

To comprehensively analyze the impact of early-exit triggering of \OurMethod on the reasoning process, we conduct a unified evaluation across four representative models, DeepSeek-R1-Distill-Qwen-7B/14B/32B and Qwen3-30B-A3B-Thinking-2507, with experimental configurations consistent with the main experiments. We report the average early-exit trigger rates across tasks and the average performance change for samples where early-exit is triggered, as depicted in Table ~\ref{tab:trigger_influnce}.

The results indicate a strong positive correlation between the early-exit trigger rate of \OurMethod and task difficulty.
Intuitively, models generally handle simpler tasks more easily, while on harder tasks they are more prone to reasoning path deviation due to error accumulation. This finding demonstrates that RPDI-EE can adaptively respond to reasoning tasks of varying difficulty. Critically, across all samples where \OurMethod triggers early-exit, model accuracy improves significantly.
This confirms that the truncated subsequent reasoning content is unproductive or redundant reasoning, and truncation effectively suppresses its generation, thereby serving a corrective function.

\input{sources/tex/trigger_influence}

\input{sources/tex/abulation}

\subsection{Ablation}
\label{sec:ablation}

In this section, we conduct experiments to verify the effectiveness of the components of RPDI and its sensitivity to hyperparameters.
We employ DeepSeek-R1-Distill-Qwen-32B as the primary model with experimental settings consistent with the main experiments.
\paragraph{Ablation on Effectiveness of the RPDI components}
We separately isolate the Local Transition Frequency (LTF), Global Transition Frequency (GTF) and Boundary-Triggered mechanism (BTM)  in RPDI to evaluate their independent contributions. 
The results are summarized in Table ~\ref{tab:ablation}.
Firstly, we remove GTF, which fixes $GTF_{i} \equiv 1$.
We find that it maintains an accuracy of 93.5\% on the \mathEasy task, but its accuracy dropped by 3.3\% on the \mathHard task.
This indicates that relying only on LTF lacks cross-task adaptability.
While still effective for simple problems, it cannot be properly calibrated for complex problems because the inherent frequency of transition tokens varies significantly with problem difficulty.
Secondly, we remove LTF by setting the sliding window size to $W=1$. This setup leads to a performance decline across all categories, validating the importance of transition token frequency in detecting reasoning path deviation.
Simply tracking the entropy change of a single token relative to the normalization baseline is insufficient for accurately detecting reasoning path deviation.
Finally, we remove the BTM and evaluate RPDI at every token generation step.
This results in an average accuracy drop of 2.8\% across all tasks, with a particularly noticeable decrease on the \mathHard task, underscoring the importance of preserving the integrity of reasoning steps.

\input{sources/tex/robostness}

\paragraph{Ablation on Hyperparameter Settings}

We conduct experiments to investigate the sensitivity of \OurMethod to different hyperparameter settings, including the window size $W$, the RPDI threshold $\lambda$, and the token budget.
Performance is systematically evaluated against the vanilla CoT baseline across various configurations.
As shown in Figure~\ref{tab:robustness}, \OurMethod consistently outperforms the vanilla CoT baseline under all tested hyperparameter settings, confirming its robustness.
In particularly, optimal accuracy is achieved with a window size $W=512$, a threshold $\lambda=2.0$, and a sufficiently large token budget, indicating stable performance within a reasonable parameter range.
We further analyze a plausible explanation for parameter influence.
The window size $W$ balances sensitivity to local noise against responsiveness to reasoning path deviations: a smaller window tends to amplify local randomness, while a huge one may delay the detection of trajectory shifts.
The threshold $\lambda$ regulates early-exit; a lower value could lead to premature truncation of the reasoning process, whereas a higher one might allow unproductive overthinking steps to persist.

\section{Conclusion}

To mitigate the degradation of reasoning performance caused by overthinking in LRLMs during long CoT reasoning, we propose \OurMethod. This method dynamically monitors internal signals of the reasoning process and adaptively triggers early-exit upon detecting reasoning path deviation.
\OurMethod is based on the insight that overthinking is frequently accompanied by the frequent occurrence of high-entropy transition tokens. By constructing the Reasoning Path Deviation Index (RPDI) to quantify the anomalous spikes of local uncertainty relative to the global baseline, it effectively resolves the issue of models falling into unproductive wandering or fragmented reasoning chains.
Extensive experiments demonstrate that compared to existing early-exit methods, \OurMethod delivers the most significant performance improvement in accuracy. 
It exhibits robust effectiveness across LRLMs of various types and scales and across tasks of varying difficulty.
These results validate the effectiveness of \OurMethod in mitigating overthinking, improving accuracy without external proxy models or probing for answers, thereby providing a new perspective for optimizing long CoT reasoning.

\section*{Impact Statement}

This paper presents work whose goal is to mitigate the performance degradation and computational redundancy caused by overthinking in LRLMs, thereby facilitating enhanced long CoT reasoning.
There are many potential societal consequences of our work, none which we feel must be specifically highlighted here.

\bibliography{example_paper}
\bibliographystyle{icml2026}
\newpage
\appendix
\onecolumn
\section{Algorithm}
\label{algorithm}
\input{sources/tex/pseudocode}
\input{sources/tex/compare_baseline}

Algorithm \ref{alg:entroscan} presents the complete execution flow of our method in pseudocode.
Through the collaborative work of the three components described above, the system can monitor the internal state of the reasoning trajectory in real-time, adaptively trigger exit upon detecting path deviation, and ultimately generate high-quality answers, thereby effectively mitigating the dual negative impact of overthinking.

\section{Implementation Details}
\label{app:experimental-details}
All experiments are implemented using the vLLM framework to ensure efficient inference and memory management.
For RPDI-EE, we set the entropy scanning window size $W$ to 512 tokens and the RPDI threshold to 2.0.
For Dynasor-CoT, we set the number of consecutive consistent results to 3 and the probing interval to 512 tokens.
Regarding context limits, we use a standard maximum length of 16,384 tokens for most models, while extending it to 32,768 tokens for the Qwen3-Thinking series to accommodate their extended reasoning chains.
The computational experiments are carried out on a cluster of 8$\times$H20 GPUs.
For all tasks, we employ greedy decoding to ensure deterministic results and use accuracy and the number of tokens generated as the primary evaluation metrics.

\section{Comparison of Early-Exit Methods}
\label{compare}
As shown in Table~\ref{tab:method-comparison}, 
By comparing various early exit strategies across six distinct dimensions, it becomes evident that our approach constitutes an online, dynamic framework.
Notably, it operates independently of answer probing or proxy models, thereby securing simultaneous gains in both predictive performance and computational efficiency.

\section{Case Study}

\begin{figure*}[p]
  \centering
   \includegraphics[width=0.8\textwidth]{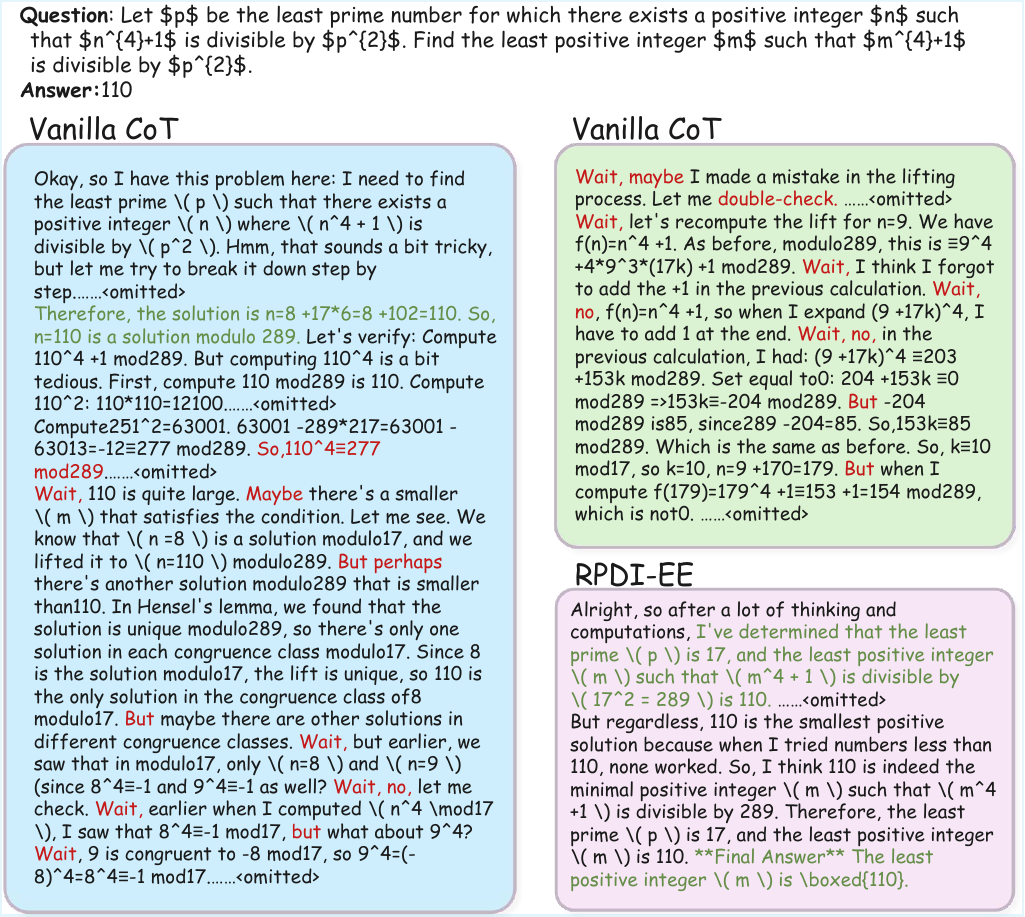}
   \caption{Comparison of generated content between \OurMethod and Vanilla on AIME24.}
  \label{case1}
\end{figure*}

\begin{figure*}[p]
  \centering
   \includegraphics[width=0.8\textwidth]{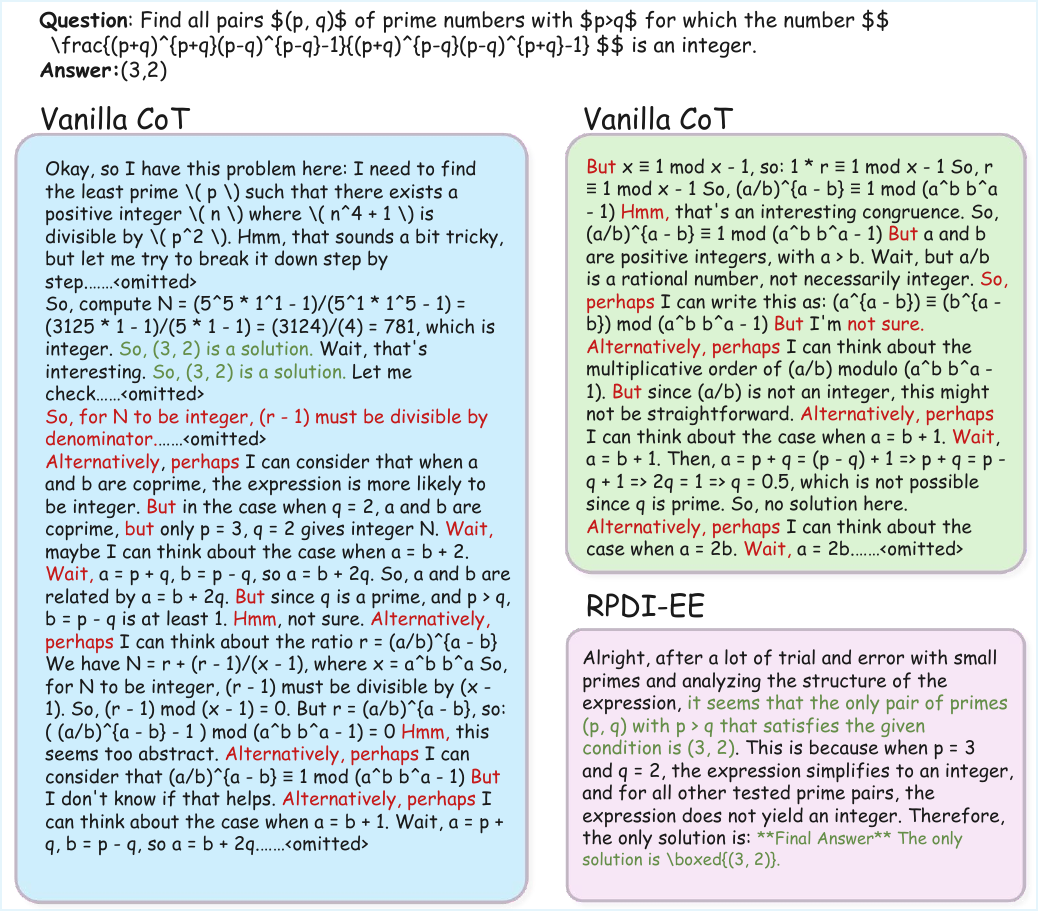}
   \caption{Comparison of generated content between \OurMethod and Vanilla on OlympiadBench.}
  \label{case2}
\end{figure*}

\begin{figure*}[p]
  \centering
   \includegraphics[width=0.8\textwidth]{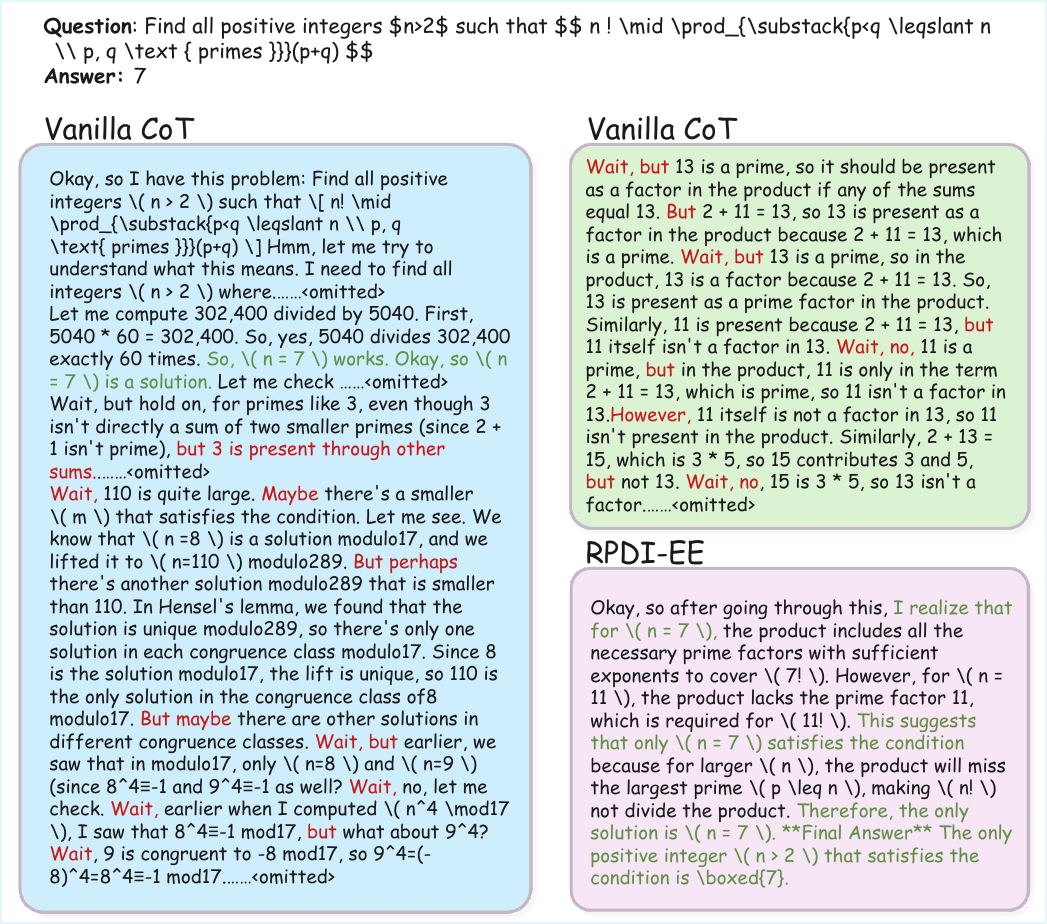}
   \caption{Comparison of generated content between \OurMethod and Vanilla on OlympiadBench.}
  \label{case3}
\end{figure*}

\begin{figure*}[p]
  \centering
   \includegraphics[width=0.8\textwidth]{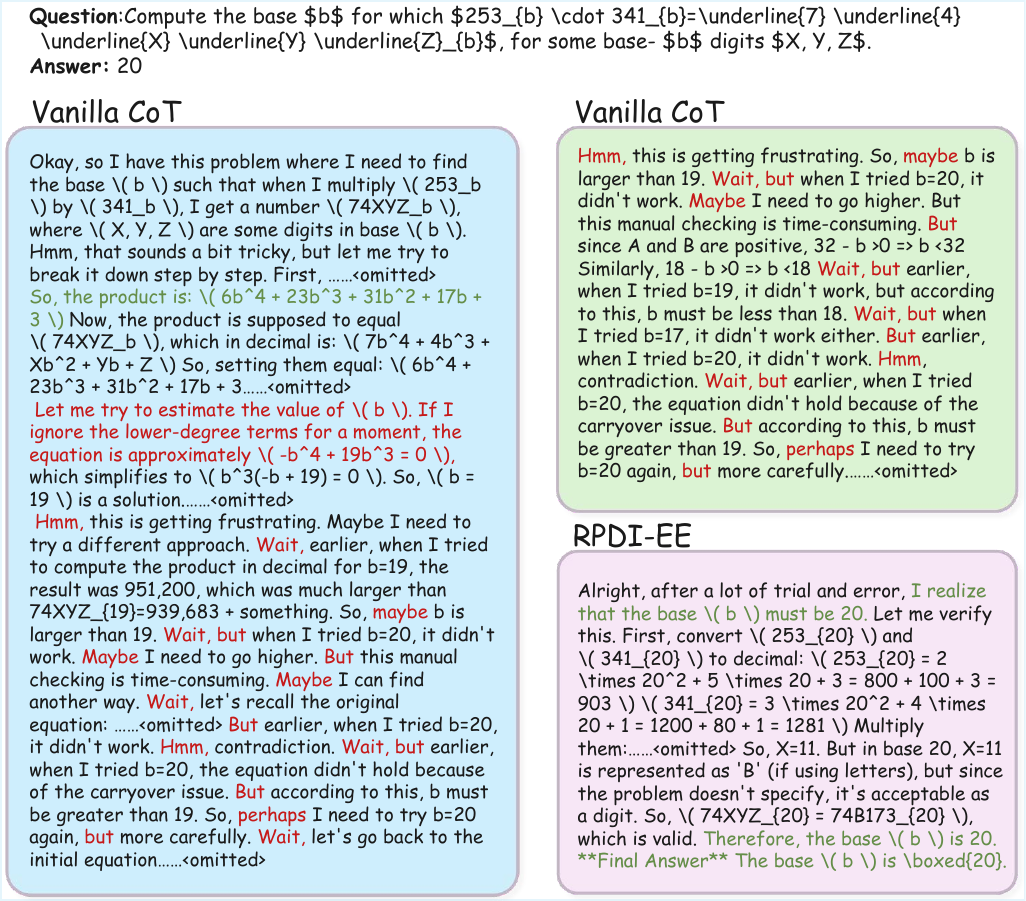}
   \caption{Comparison of generated content between \OurMethod and Vanilla on OlympiadBench.}
  \label{case4}
\end{figure*}

In Figure ~\ref{case1}, we present result a example on the AIME2024 dataset to intuitively demonstrate the effectiveness of \OurMethod.
Due to space constraints, where the full presentation of a sample typically spans several pages, we have omitted some intermediate steps to highlight the key information.
Prior to triggering the early-exit, \OurMethod follows the same reasoning trajectory as vanilla CoT, as shown in the blue box. At this stage, the model initially achieves the correct result, but subsequently generates a calculation error that triggers a logical contradiction and leads to reasoning path deviation.
This state is marked by the frequent occurrence of high-entropy transition tokens (e.g., “Wait”, “But”), indicating that the model is producing fragmented reasoning chains and falling into unproductive wandering.
The green box and the pink box illustrate two distinct subsequent results: while vanilla CoT remains trapped in a redundant verification loop, our method detects the reasoning path deviation, allowing the redundant inference to be effectively suppressed.
As shown in the pink box, which displays the results after truncation, the model achieves self-rectification and ultimately generates the correct final answer.
More examples are provided in Figures ~\ref{case2}, ~\ref{case3}, and ~\ref{case4}; notably, in the case of Figure ~\ref{case4}, although the model fails to compute the correct result before triggering early-exit , it has already obtained sufficient intermediate reasoning steps, and \OurMethod prevents further confusion caused by error accumulation.


\end{document}

%% file: sources/tex/compare.tex
\begin{figure}[t]
  \centering
  \includegraphics[width=0.45\textwidth]{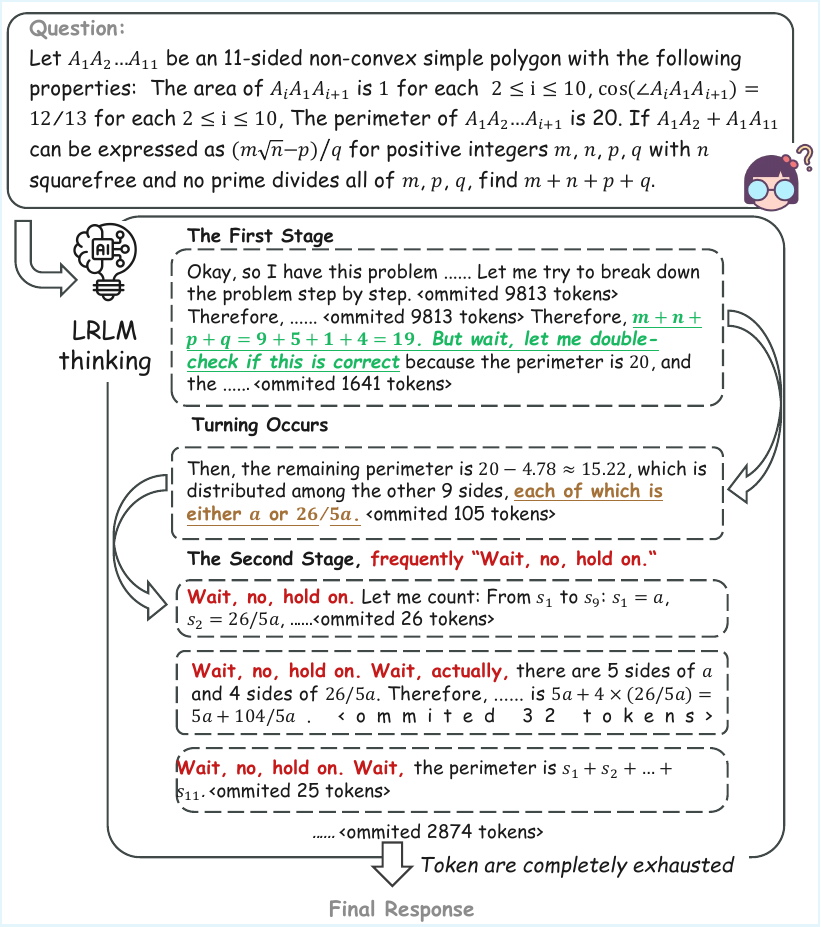}
  \caption{Reasoning trajectory dissection revealing the ``overthinking" trap. The model initially achieves the correct result but fails due to a flawed verification loop. The frequent emergence of transition tokens (\textit{e.g.}, ``Wait," ``But") serves as a key indicator of reasoning deviation.}
  \label{fig:intro_case}
\end{figure}

%% file: sources/tex/high_entropy_word_weight.tex
\begin{figure}[!t]
  \centering
  \includegraphics[width=0.48\textwidth]{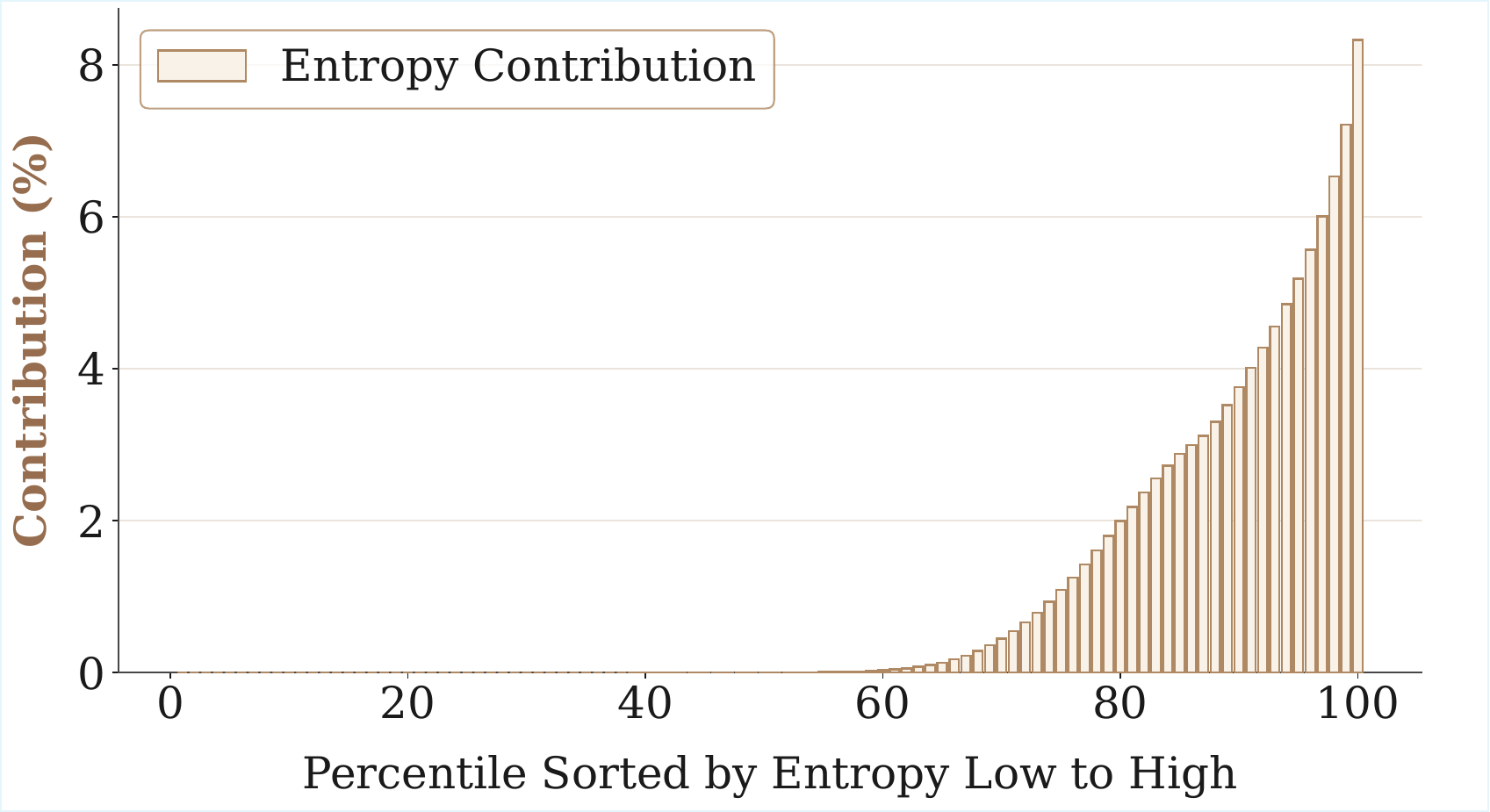}
  \caption{\textbf{Token Entropy Contribution Distribution.} This figure illustrates the distribution of $45.9$ million tokens, sorted by their entropy values from low to high. The tokens are divided into 100 percentile bins, with the y-axis representing the percentage contribution of the total entropy within each bin relative to the aggregate entropy of all tokens. 
  }
  \label{fig:preliminary1}
\end{figure}

%% file: sources/tex/high_entropy_word.tex
\begin{figure}[!t]
  \centering
  \includegraphics[width=0.48\textwidth]{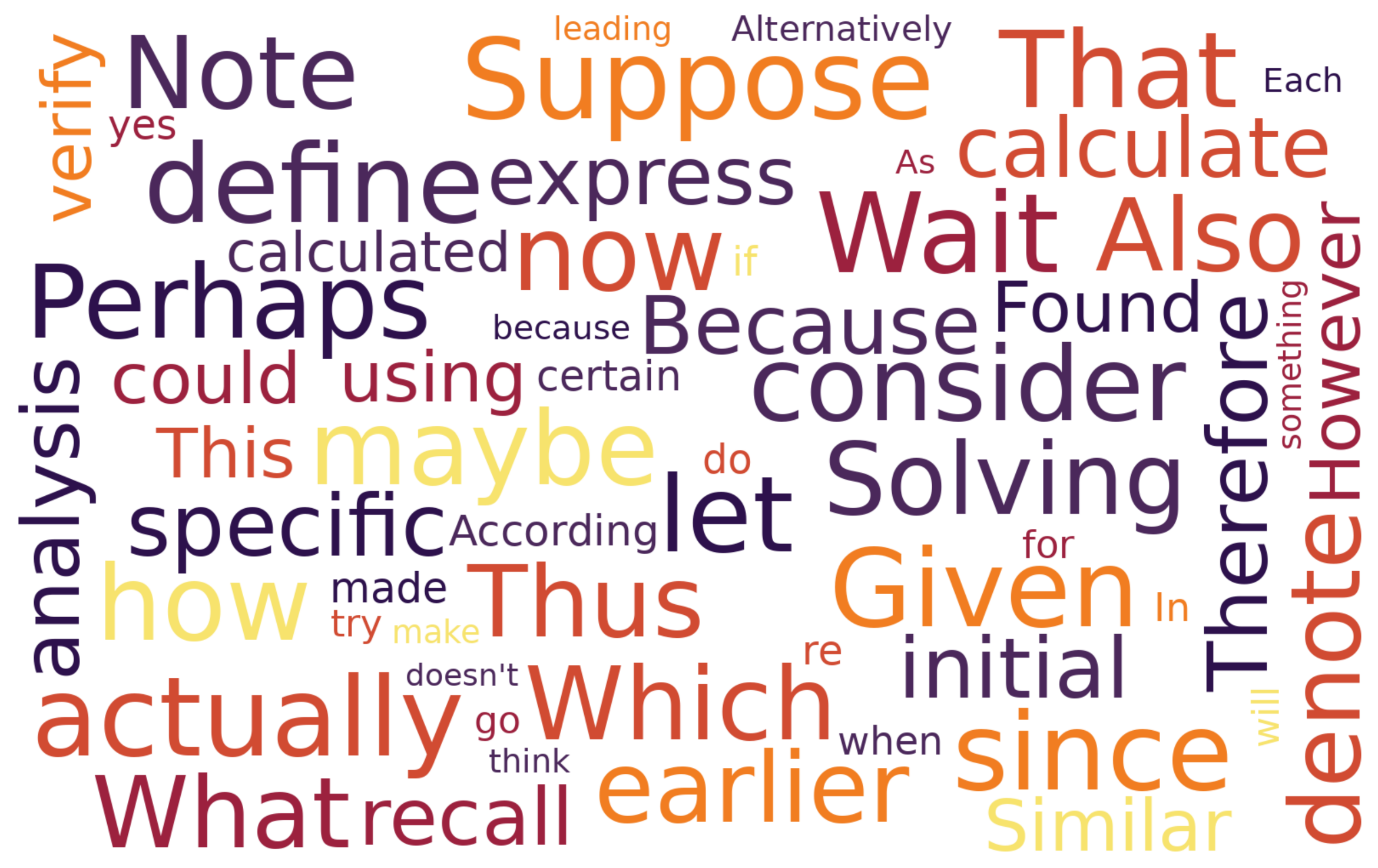}
  \caption{\textbf{Visualization of High-Frequency Tokens among Top Entropy Contributors.} This figure highlights the most frequent tokens within the $20\%$ of tokens that contributed most to the average entropy. 
  Only tokens that constitute complete English words are retained.
  }
  \label{fig:preliminary2}
\end{figure}    

%% file: sources/tex/method.tex
\begin{figure*}[!ht]
  \centering
   \includegraphics[width=\textwidth]{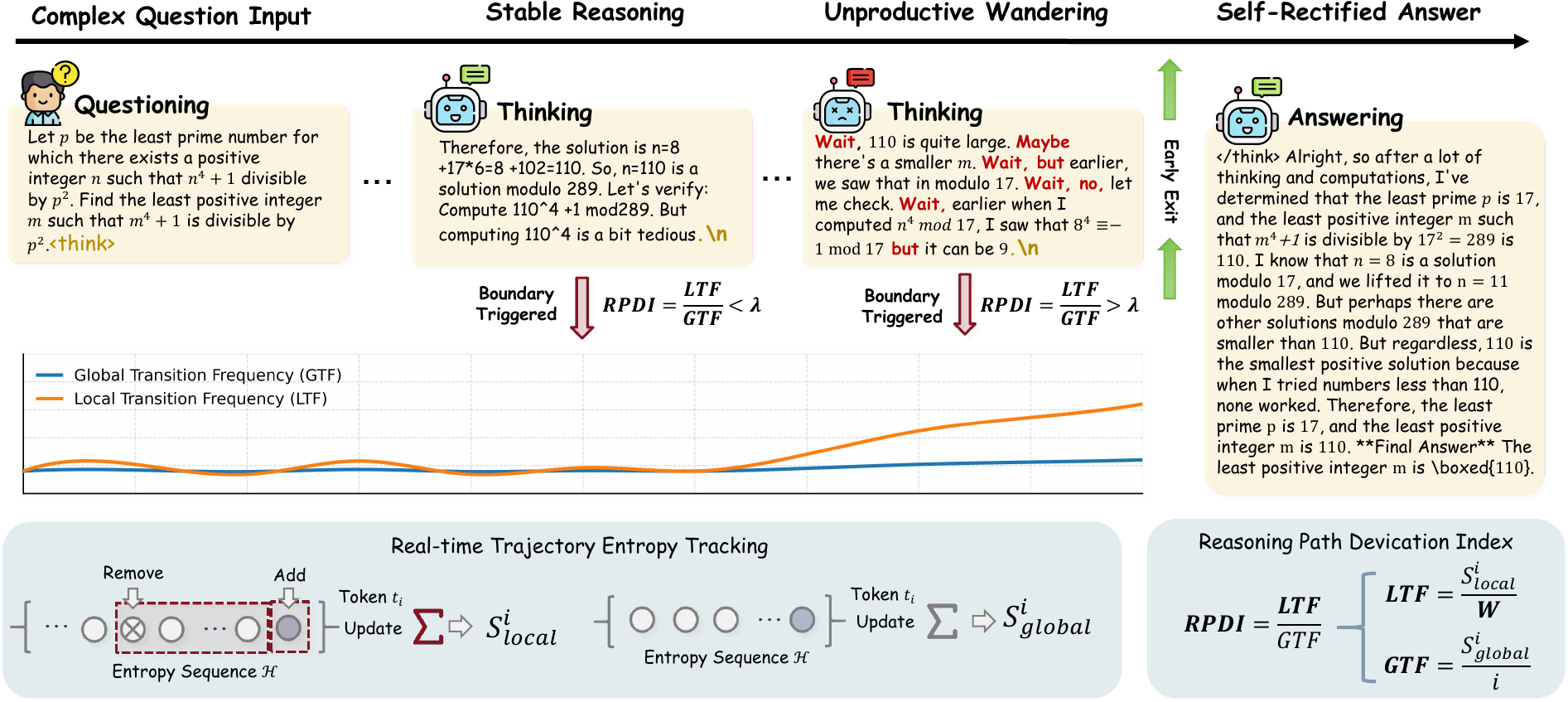}
  \caption{\textbf{Overview of RPDI-EE.} 
  RPDI-EE performs continuous entropy scanning during CoT generation. For each new token, it measures local uncertainty via the Local Reasoning Density (LRD), defined as the average token entropy in a sliding window, and compares it to the global reasoning stability reflected by the Global Reasoning Baseline (GRB). 
  }
  \label{fig:method}
\end{figure*}

%% file: sources/tex/main_result.tex
\begin{table*}[!t]
\caption{Performance comparison. 
Subscripts in all Acc. columns denote the accuracy change relative to the Vanilla CoT.
Limited by computing resources and costs, \OurMethod is only compared with Vanilla CoT and the most competitive DEER across all eight LRLMs.
}
\centering
\small
\begin{tabular}{llrlrlrlr}
\toprule
& \multicolumn{2}{c}{\textbf{\mathEasy}} 
& \multicolumn{2}{c}{\textbf{\mathHard}} 
& \multicolumn{2}{c}{\textbf{\scientificBC}} 
& \multicolumn{2}{c}{\textbf{\textsc{Average}}} \\ 
\cmidrule(lr){2-3} \cmidrule(lr){4-5} \cmidrule(lr){6-7} \cmidrule(lr){8-9}

\multirow{-2}{*}{\textbf{\textsc{Method}}} 
& Acc.$\uparrow$ & Len.$\downarrow$ & Acc.$\uparrow$ & Len.$\downarrow$ & Acc.$\uparrow$ & Len.$\downarrow$ & Acc.$\uparrow$ & Len.$\downarrow$ \\ 

\midrule

\rowcolor[HTML]{E7E6E6} 
\textbf{DeepSeek-R1-Distill-Qwen-7B} & 87.9 & 3801 & 32.5 & 10395 & 32.3 & 10319 & 56.2 & 7558 \\
\quad + NoThinking & 77.6$_{\color{red}{-10.3}}$ & 912 & 25.2$_{\color{red}{-7.3}}$ & 3277 & 31.3$_{\color{red}{-1.0}}$ & 841 & 48.5$_{\color{red}{-7.7}}$ & 1915 \\
\quad + ThinkLess & 76.8$_{\color{red}{-11.1}}$ & 758 & 24.7$_{\color{red}{-7.8}}$ & 3736 & 31.8$_{\color{red}{-0.5}}$ & 1175 & 48.0$_{\color{red}{-8.2}}$ & 2094 \\
\quad + Dynasor-CoT & 73.8$_{\color{red}{-14.1}}$ & 1673 & 32.4$_{\color{red}{-0.1}}$ & 3366 & \textbf{45.5}$_{\colorgreen{+13.2}}$ & 1876 & 52.0$_{\color{red}{-4.2}}$ & 2428 \\
\quad + DEER & 85.4$_{\color{red}{-2.5}}$ & 2733 & 45.0$_{\colorgreen{+12.5}}$ & 9005 & 28.3$_{\color{red}{-4.0}}$ & 9433 & 59.9$_{\colorgreen{+3.7}}$ & 6378 \\
\quad + \textbf{\OurMethod} & \textbf{88.0}$_{\colorgreen{+0.1}}$ & 3825 & \textbf{47.9}$_{\colorgreen{+15.4}}$ & 9700 & 24.8$_{\color{red}{-7.5}}$ & 9727 & \textbf{61.8}$_{\colorgreen{+5.6}}$ & 7186 \\

\midrule

\rowcolor[HTML]{E7E6E6} 
\textbf{DeepSeek-R1-Distill-Qwen-14B} & 90.0 & 3549 & 53.2 & 9464 & 50.5 & 6963 & 68.6 & 6572 \\
\quad + NoThinking & 80.5$_{\color{red}{-9.5}}$ & 1187 & 31.1$_{\color{red}{-22.1}}$ & 5641 & 44.4$_{\color{red}{-6.1}}$ & 1309 & 54.2$_{\color{red}{-14.4}}$ & 3113 \\
\quad + ThinkLess & 63.8$_{\color{red}{-26.2}}$ & 1394 & 27.7$_{\color{red}{-25.5}}$ & 5937 & 34.9$_{\color{red}{-15.6}}$ & 1748 & 44.2$_{\color{red}{-24.4}}$ & 3391 \\
\quad + Dynasor-CoT & 76.4$_{\color{red}{-13.6}}$ & 1535 & 30.8$_{\color{red}{-22.4}}$ & 2985 & 48.5$_{\color{red}{-2.0}}$ & 1686 & 52.9$_{\color{red}{-15.7}}$ & 2178 \\
\quad + DEER & 90.9$_{\colorgreen{+0.9}}$ & 2665 & 53.2 & 9159 & 52.0$_{\colorgreen{+1.5}}$ & 6575 & 69.2$_{\colorgreen{+0.6}}$ & 6007 \\
\quad + \textbf{\OurMethod} & \textbf{93.5}$_{\colorgreen{+3.5}}$ & 3426 & \textbf{59.2}$_{\colorgreen{+6.0}}$ & 9282 & \textbf{56.1}$_{\colorgreen{+5.6}}$ & 6575 & \textbf{73.4}$_{\colorgreen{+4.8}}$ & 6386 \\

\midrule

\rowcolor[HTML]{E7E6E6} 
\textbf{DeepSeek-R1-Distill-Qwen-32B} & 88.7 & 3753 & 50.4 & 9685 & 56.1 & 7386 & 67.6 & 6814 \\
\quad + NoThinking & 82.1$_{\color{red}{-6.6}}$ & 797 & 40.3$_{\color{red}{-10.1}}$ & 6531 & 51.0$_{\color{red}{-5.1}}$ & 2249 & 59.7$_{\color{red}{-7.9}}$ & 3462 \\
\quad + ThinkLess & 77.1$_{\color{red}{-11.6}}$ & 485 & 29.2$_{\color{red}{-21.2}}$ & 2796 & 53.5$_{\color{red}{-2.6}}$ & 4365 & 53.2$_{\color{red}{-14.4}}$ & 2030 \\
\quad + Dynasor-CoT & 75.1$_{\color{red}{-13.6}}$ & 1390 & 34.0$_{\color{red}{-16.4}}$ & 2392 & 57.6$_{\colorgreen{+1.5}}$ & 1655 & 55.0$_{\color{red}{-12.6}}$ & 1857 \\
\quad + DEER & 90.9$_{\colorgreen{+2.2}}$ & 2757 & \textbf{61.0}$_{\colorgreen{+10.6}}$ & 8637 & 52.0$_{\color{red}{-4.1}}$ & 7134 & 72.5$_{\colorgreen{+4.9}}$ & 5902 \\
\quad + \textbf{\OurMethod} & \textbf{93.5}$_{\colorgreen{+4.8}}$ & 3210 & \textbf{61.0}$_{\colorgreen{+10.6}}$ & 9119 & \textbf{64.7}$_{\colorgreen{+8.6}}$ & 6972 & \textbf{75.5}$_{\colorgreen{+7.9}}$ & 6280 \\

\midrule

\rowcolor[HTML]{E7E6E6} 
\textbf{Qwen3-30B-A3B-Thinking-2507} & 96.7 & 5178 & 82.1 & 15621 & 74.8 & 7394 & 87.3 & 9970 \\
\quad + NoThinking & 92.2$_{\color{red}{-4.5}}$ & 2820 & 61.7$_{\color{red}{-20.4}}$ & 6923 & 71.2$_{\color{red}{-3.6}}$ & 6267 & 76.2$_{\color{red}{-11.1}}$ & 5071 \\
\quad + ThinkLess & 96.9$_{\colorgreen{+0.2}}$ & 4409 & 81.8$_{\color{red}{-0.3}}$ & 14418 & 74.8 & 6841 & 87.2$_{\color{red}{-0.1}}$ & 9046 \\
\quad + Dynasor-CoT & 74.6$_{\color{red}{-22.1}}$ & 1504 & 28.6$_{\color{red}{-53.5}}$ & 2525 & 59.1$_{\color{red}{-15.7}}$ & 1735 & 52.7$_{\color{red}{-34.6}}$ & 1975 \\
\quad + DEER & 97.2$_{\colorgreen{+0.5}}$ & 4477 & 80.5$_{\color{red}{-1.6}}$ & 15285 & 74.8 & 7167 & 86.8$_{\color{red}{-0.5}}$ & 9493 \\
\quad + \textbf{\OurMethod} & \textbf{97.4}$_{\colorgreen{+0.7}}$ & 5039 & \textbf{84.6}$_{\colorgreen{+2.5}}$ & 14927 & \textbf{77.8}$_{\colorgreen{+3.0}}$ & 7393 & \textbf{89.1}$_{\colorgreen{+1.8}}$ & 9613 \\

\midrule

\rowcolor[HTML]{E7E6E6} 
\textbf{DeepSeek-R1-Distill-Qwen-1.5B} & 65.7 & 5615 & 25.7 & 12646 & 6.1 & 13386 & 40.0 & 9738 \\
\quad + DEER & 64.3$_{\color{red}{-1.4}}$ & 3202 & 24.7$_{\color{red}{-1.0}}$ & 10155 & 4.6$_{\color{red}{-1.5}}$ & 12685 & 38.8$_{\color{red}{-1.2}}$ & 7536 \\
\quad + \textbf{\OurMethod} & \textbf{69.4}$_{\colorgreen{+3.7}}$ & 5177 & \textbf{29.2}$_{\colorgreen{+3.5}}$ & 11394 & \textbf{8.1}$_{\colorgreen{+2.0}}$ & 12823 & \textbf{43.4}$_{\colorgreen{+3.4}}$ & 8934 \\

\midrule

\rowcolor[HTML]{E7E6E6} 
\textbf{DeepSeek-R1-Distill-Llama-8B} & 70.8 & 5036 & 32.5 & 10876 & 21.2 & 9604 & 47.3 & 8192 \\
\quad + DEER & 73.6$_{\colorgreen{+2.8}}$ & 3383 & 33.1$_{\colorgreen{+0.6}}$ & 10575 & \textbf{34.3}$_{\colorgreen{+13.1}}$ & 9515 & 50.6$_{\colorgreen{+3.3}}$ & 7341 \\
\quad + \textbf{\OurMethod} & \textbf{82.8}$_{\colorgreen{+12.0}}$ & 4271 & \textbf{32.8}$_{\colorgreen{+0.3}}$ & 10905 & 25.3$_{\colorgreen{+4.1}}$ & 9769 & \textbf{53.2}$_{\colorgreen{+5.9}}$ & 7900 \\

\midrule

\rowcolor[HTML]{E7E6E6} 
\textbf{DeepSeek-R1-Distill-Llama-70B} & 89.0 & 3639 & 51.4 & 9870 & 60.1 & 5538 & 68.8 & 6581 \\
\quad + DEER & \textbf{91.7}$_{\colorgreen{+2.7}}$ & 2200 & 46.4$_{\color{red}{-5.0}}$ & 8528 & 62.1$_{\colorgreen{+2.0}}$ & 5659 & 68.0$_{\color{red}{-0.8}}$ & 5406 \\
\quad + \textbf{\OurMethod} & 89.4$_{\colorgreen{+0.4}}$ & 3876 & \textbf{55.8}$_{\colorgreen{+4.4}}$ & 9386 & \textbf{64.7}$_{\colorgreen{+4.6}}$ & 5940 & \textbf{71.5}$_{\colorgreen{+2.7}}$ & 6532 \\

\midrule

\rowcolor[HTML]{E7E6E6} 
\textbf{Qwen3-235B-Thinking} & 97.7 & 5805 & 79.8 & 17185 & 80.8 & 8893 & 87.6 & 11123 \\
\quad + DEER & 97.7 & 4591 & 82.5$_{\colorgreen{+2.7}}$ & 16332 & \textbf{82.8}$_{\colorgreen{+2.0}}$ & 9005 & 89.1$_{\colorgreen{+1.5}}$ & 10253 \\
\quad + \textbf{\OurMethod} & \textbf{97.8}$_{\colorgreen{+0.1}}$ & 5869 & \textbf{84.8}$_{\colorgreen{+5.0}}$ & 16642 & 80.3$_{\color{red}{-0.5}}$ & 8910 & \textbf{89.7}$_{\colorgreen{+2.1}}$ & 10921 \\

\bottomrule
\end{tabular}
\label{tab:comprehensive_comparison_final}
\end{table*}

%% file: sources/tex/trigger_influence.tex
\begin{figure}[!t]
  \centering
  \includegraphics[width=0.48\textwidth]{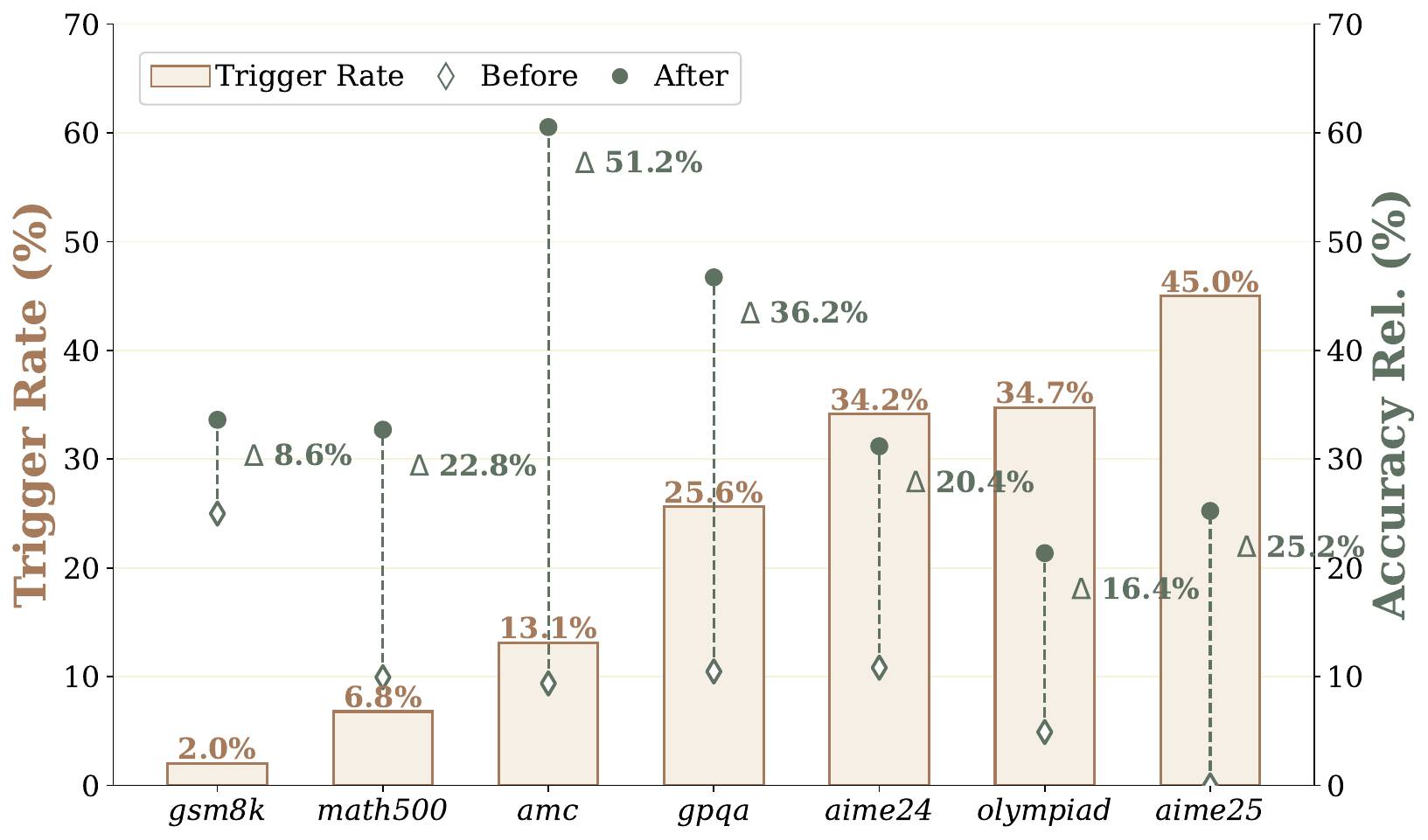}
  \caption{Triggering Rates and Corrective Performance.}
  \label{tab:trigger_influnce}
\end{figure}

%% file: sources/tex/abulation.tex
\begin{table}[!t]
\caption{Effectiveness Ablation of the RPDI components.}
\centering
\small
\begin{tabular}{lcccc}
\toprule
\multirow{2}{*}{\textbf{\textsc{Method}}} & \textbf{\textsc{Math}} & \textbf{\textsc{Math}} & \multirow{2}{*}{\textbf{\textsc{Scientific}}} & \multirow{2}{*}{\textbf{\textsc{{Average}}}} \\
 & \textbf{\textsc{Easy}} & \textbf{\textsc{Hard}} & & \\
\midrule
\textbf{Vanilla CoT} & 91.6 & 52.9 & 52.0 & 67.6 \\
\textbf{\OurMethod} & 93.5 & \textbf{61.0} & 64.7 & \textbf{75.5} \\
\quad w/o GTF & 93.5 & 57.7 & 64.7 & 74.0 \\
\quad w/o LTF & 90.2 & 56.4 & 62.1 & 71.7 \\
\quad w/o BTM & \textbf{93.6} & 53.5 & \textbf{67.7} & 72.7 \\
\bottomrule
\end{tabular}
\label{tab:ablation}
\end{table}

%% file: sources/tex/robostness.tex
\begin{figure}[!t]
  \centering
  \includegraphics[width=0.48\textwidth]{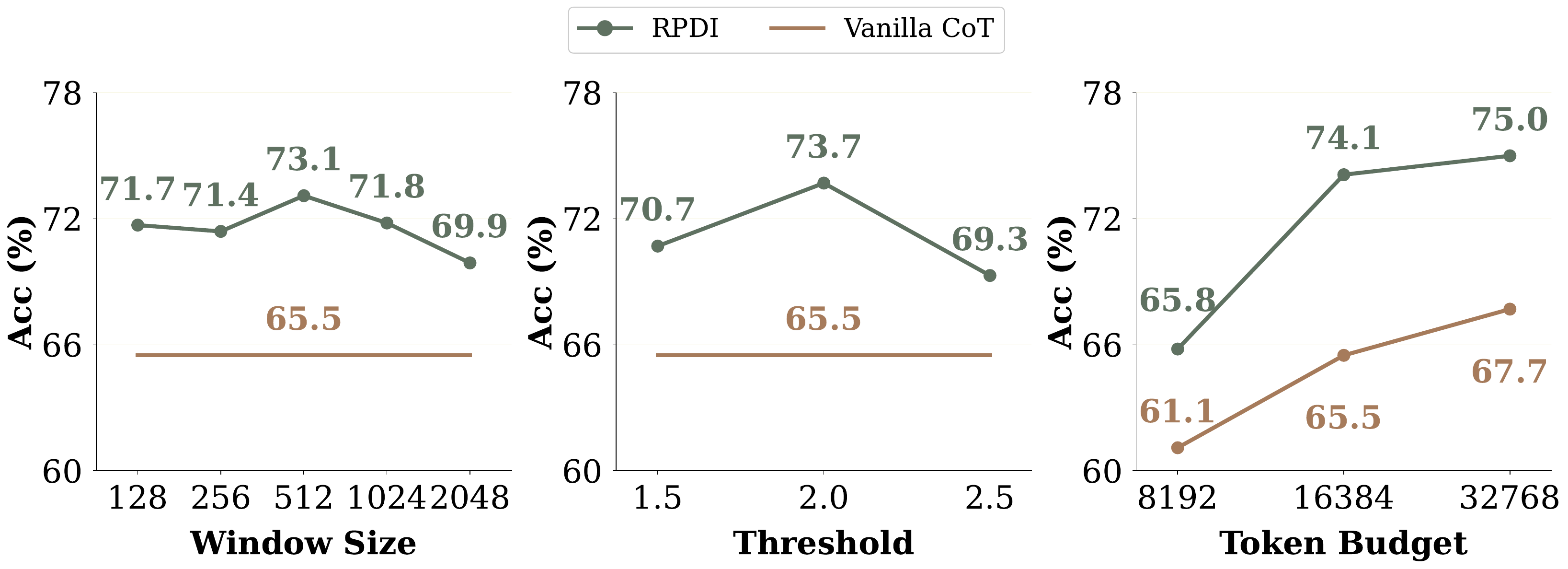}
  \caption{Hyperparameter Ablation of \OurMethod. 
  }
  \label{tab:robustness}
\end{figure}

%% file: sources/tex/pseudocode.tex
\begin{algorithm}[t]
\caption{Our Method}
\label{alg:entroscan}
\small
\begin{algorithmic}[1]
    \REQUIRE 
        Model $M$, prompt $P$, token budget $L_{\text{max}}$, 
        local content size $W$, threshold $\lambda$, boundary set $\mathcal{B}$

    \ENSURE 
        Final answer $A$

    \STATE \textcolor[HTML]{6A9955}{/* Initialize trajectory and entropy buffers */}
    \STATE $R \gets \emptyset, \mathcal{H} \gets [\,], S_{\text{global}} \gets 0.0, S_{\text{local}} \gets 0.0$
    \STATE $i \gets 0, \text{early\_exit} \gets \text{False}$

    \STATE \textcolor[HTML]{6A9955}{/* Introspective loop with real-time scanning */}
    \WHILE{\textbf{not} \text{early\_exit}}
        \STATE $i \gets i + 1$
        \STATE Sample $t_i \sim M(P \oplus R)$, compute entropy $H(t_i)$
        \STATE $R \gets R \oplus t_i,  \mathcal{H}.\text{append}(H(t_i))$
        
        \STATE \textcolor[HTML]{6A9955}{/* Incremental entropy tracking */}
        \STATE $S_{\text{global}} \gets S_{\text{global}} + H(t_i), S_{\text{local}} \gets S_{\text{local}} + H(t_i)$
        
        \STATE \textcolor[HTML]{6A9955}{/* Local context update */}
        \IF{$i > W$}
            \STATE $S_{\text{local}} \gets S_{\text{local}} - \mathcal{H}[i-W]$
        \ENDIF

        \STATE \textcolor[HTML]{6A9955}{/* Early exit via RPDI detection */}
        \IF{$i \ge W$ \AND $t_i \in \mathcal{B}$} 
            \STATE $\text{GTF} \gets S_{\text{global}} / i, \text{LTF} \gets S_{\text{local}} / W$ 
            \STATE $\text{RPDI} \gets \text{LTF} / \text{GTF}$ 
            \STATE $\text{early\_exit} \gets \text{RPDI} > \lambda$
        \ENDIF

    \ENDWHILE

    \STATE \textcolor[HTML]{6A9955}{/* Transition to answer synthesis phase */}
    \STATE $R \gets R \oplus \text{\texttt{</think>}}$
    \STATE Sample $A \sim M(P \oplus R)$ within budget $L_{\text{max}}-i$
    
    \STATE \textbf{Return} $A$
\end{algorithmic}
\end{algorithm}

%% file: sources/tex/compare_baseline.tex
\begin{table*}[t]
\caption{Feature-level comparison between RPDI and existing early-exit frameworks. We evaluate methods across six critical dimensions: (1) \textbf{Probing-free}: avoids intermediate answer generation to maintain a continuous reasoning flow; (2) \textbf{Proxy-Model-free}: operates without auxiliary verifiers or additional training; (3) \textbf{Dynamic Method}: adaptively adjusts the reasoning length based on problem complexity; (4) \textbf{Efficiency Gain}: provides measurable reduction in latency compared to vanilla CoT; (5) \textbf{Performance Gain}: achieves accuracy improvements by mitigating overthinking while safeguarding against over-truncation; (6) \textbf{Real-time}: functions during the generation process rather than via offline post-processing.}
\centering
\small
\begin{tabular}{@{}lcccccc@{}}
\toprule
\textbf{Method} & 
\rotatebox{45}{Probing-free} & 
\rotatebox{45}{Proxy-Model-free} & 
\rotatebox{45}{Dynamic Method} & 
\rotatebox{45}{Efficiency Gain} & 
\rotatebox{45}{Performance Gain} & 
\rotatebox{45}{Real-time} \\
\midrule
\textbf{RPDI (Ours)} & \cmark & \cmark & \cmark & \cmark & \cmark & \cmark \\
\midrule
\rowcolor[gray]{0.95} \multicolumn{7}{@{}l}{\textit{Answer-Probing Based}} \\
Deer \citep{yang2025dynamic} & \xmark & \cmark & \cmark & \cmark & \cmark & \cmark \\
Adaptive Think \citep{yong2025think} & \xmark & \cmark & \cmark & \cmark & \cmark & \xmark \\
Dynasor-CoT \citep{fu2025reasoning} & \xmark & \cmark & \cmark & \cmark & \xmark & \cmark \\
\midrule
\rowcolor[gray]{0.95} \multicolumn{7}{@{}l}{\textit{Proxy-Model Based}} \\
SpecExit \citep{yang2025specexit} & \cmark & \xmark & \cmark & \cmark & \xmark & \cmark \\
Flashthink \citep{jiang2025flashthink} & \cmark & \xmark & \cmark & \cmark & \xmark & \cmark \\
Zhang et al. \citep{zhang2025reasoning} & \cmark & \xmark & \cmark & \cmark & \xmark & \cmark \\
LUNX \citep{akgul2025lynx} & \xmark & \xmark & \cmark & \cmark & \cmark & \cmark \\
\midrule
\rowcolor[gray]{0.95} \multicolumn{7}{@{}l}{\textit{Fixed LenghBased}} \\
Token Budget \citep{muennighoff2025s1} & \cmark & \cmark & \xmark & \cmark & \xmark & \cmark \\
ThinkLess \citep{li2025thinkless} & \cmark & \cmark & \xmark & \cmark & \xmark & \cmark \\
NoThinking \citep{ma2025reasoning} & \cmark & \cmark & \xmark & \cmark & \xmark & \cmark \\
\bottomrule
\end{tabular}
\label{tab:method-comparison}
\end{table*}